\documentclass[conference]{IEEEtran}
\IEEEoverridecommandlockouts

\usepackage{cite}
\usepackage{amsmath,amssymb,amsfonts}
\usepackage{algorithmic}
\usepackage{algorithm}
\usepackage{graphicx}
\usepackage{textcomp}
\usepackage{xcolor}
\usepackage{url}
\usepackage{float}
\usepackage{placeins}
\usepackage{booktabs}
\usepackage[belowskip=-5pt,aboveskip=3pt]{caption}
\usepackage{hyperref}
\def\BibTeX{{\rm B\kern-.05em{\sc i\kern-.025em b}\kern-.08em
    T\kern-.1667em\lower.7ex\hbox{E}\kern-.125emX}}
\begin{document}

\title{Hierarchical and Query-Aware Visual Token Pruning for High-Resolution LVLMs\\}


\author{\IEEEauthorblockN{Luther Wei\textsuperscript{1}, Ling Zou\textsuperscript{1}, Kang-Yang Huang\textsuperscript{1}, Yu-Syuan Xu\textsuperscript{1, 2}, and Wen-Huang Cheng\textsuperscript{1}}
\IEEEauthorblockA{
\textsuperscript{1}\textit{Department of Computer Science and Information Engineering, National Taiwan University}, Taipei, Taiwan \\
\textsuperscript{2}\textit{MediaTek Inc.}, Hsinchu, Taiwan \\
weiluther@gmail.com, \{zouling, huangkangyang\}@cmlab.csie.ntu.edu.tw, \\
yu-syuan.xu@mediatek.com, wenhuang@csie.ntu.edu.tw}

}

\maketitle


\begin{abstract}
High-resolution Large Vision-Language Models (LVLMs) preserve critical, fine-grained visual details by encoding a global thumbnail alongside multiple local crops. However, the resulting thousands of vision tokens introduce severe computational bottlenecks during Large Language Model (LLM) inference. Existing token pruning methods rely primarily on query-agnostic visual saliency, while current query-aware approaches either operate at a coarse crop-level resolution or compute token relevance too late, after vision tokens have already entered the heavy LLM layers. 
To address this, we propose a training-free, hierarchical token pruning framework that utilizes query guidance prior to executing any LLM blocks. Our top-down hierarchy first operates at the instance-level, dynamically scaling the vision token retention ratio of each sample based on its visual complexity. Next, at the crop-level, we distribute this sample budget across local crops using sentence-level cross-modal similarity. Finally, at the token-level, we exploit projected vision tokens alongside textual query token embeddings in the shared space to measure fine-grained alignment, applying a text-debiasing normalization to produce high-fidelity query scores. These query-guided budgets and scores are fused with local and global visual saliency metrics to select the most critical vision tokens.
Extensive experiments on LLaVA-NeXT-7B demonstrate that our plug-and-play framework retains 98.0\% and 94.8\% of unpruned performance at 25\% and 10\% vision token retention, respectively. Notably, at 25\% retention, our method delivers a 2.0$\times$ inference throughput speedup and even surpasses the unpruned baseline on POPE by 0.7 points.
\end{abstract}

\section{Introduction}

Large Vision-Language Models (LVLMs)~\cite{liu2023llava,dai2023instructblip,wang2024qwen2vl} provide a powerful general interface for multimodal reasoning, yet applications like document understanding, chart analysis, and small-object recognition demand high-resolution inputs to preserve critical, fine-grained visual details. To accommodate this, modern high-resolution LVLMs~\cite{liu2024llavanext} encode a global thumbnail alongside multiple local crops, projecting and concatenating their features into the Large Language Model (LLM) input space. However, this multi-crop strategy drastically increases the vision token count, often exceeding several thousand tokens per image. Because the computational complexity of the LLM's self-attention mechanism scales quadratically with sequence length~\cite{vaswani2017attention}, this token explosion severely bottlenecks inference speed, inflates memory consumption, and hinders deployment on resource-constrained devices.


Existing token-reduction approaches are broadly divided into training-based methods, which learn compact token projectors but require expensive retraining~\cite{cha2024honeybee,li2024tokenpacker}, and training-free methods, which prune or merge vision tokens at inference time~\cite{chen2024fastv,zhang2024fastervlm}. While training-free methods offer plug-and-play flexibility, most are designed for single-grid images and fail to handle the global-thumbnail-and-local-crop layout of high-resolution LVLMs. Recent frameworks like HiRED~\cite{arif2025hired} and GlobalCom$^2$~\cite{liu2026globalcom2} address this architecture via a two-stage pipeline: a budget allocation stage that determines how many tokens each crop receives, and a token selection stage that decides which vision tokens to preserve within that crop.

\begin{figure}[!t]
\centering
\includegraphics[width=\columnwidth]{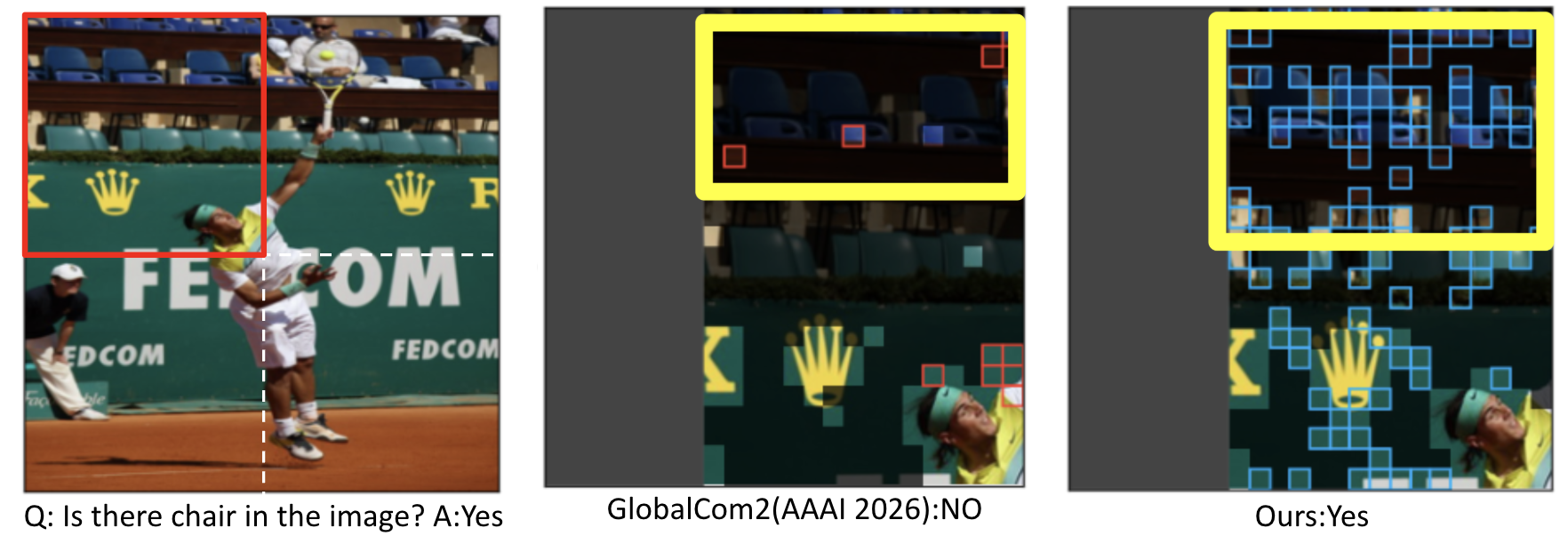}
\caption{Pruning masks on a POPE sample at a 25\% retention ratio. GlobalCom$^2$ uses an image-only mask that is unchanged by the query; for the chair question shown here, it removes most chair evidence before the LLM and answers incorrectly. Our query-aware mask retains the relevant chair tokens and answers correctly.}
\label{fig:pruningmask}
\end{figure}

\begin{figure}[!t]
\centering
\includegraphics[width=\columnwidth]{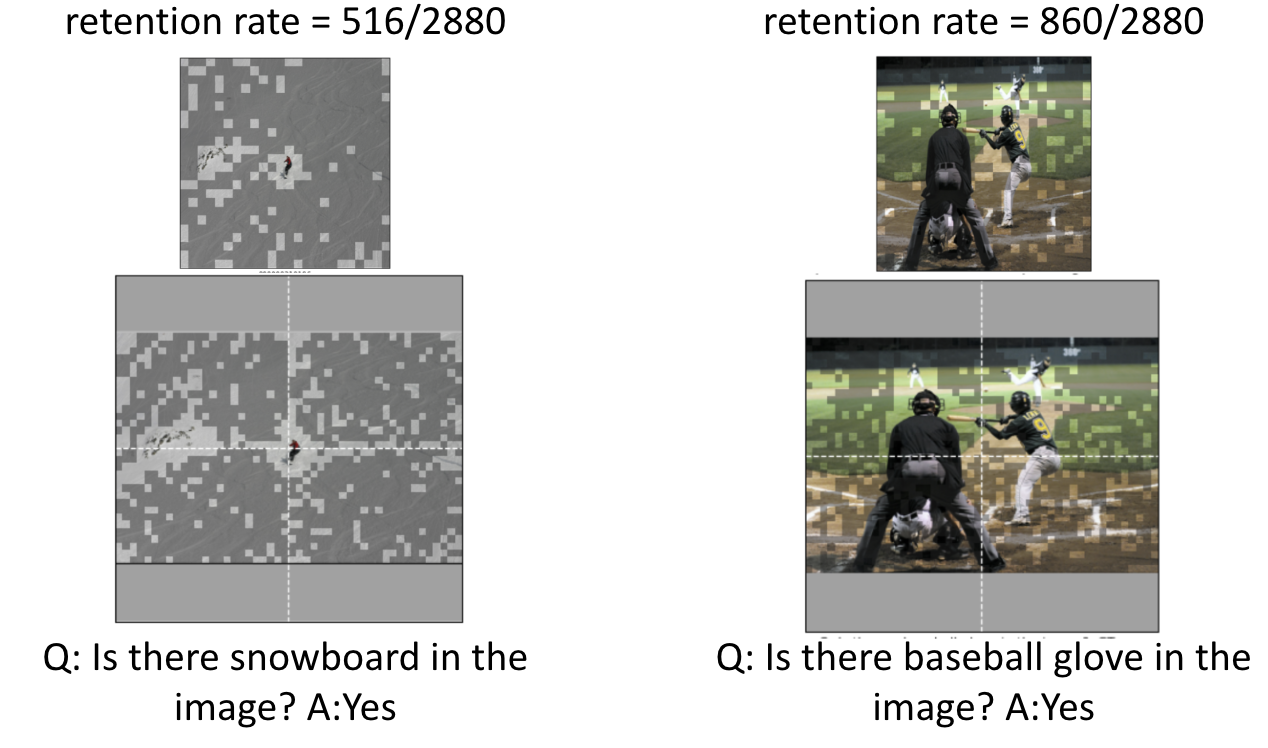}
\caption{Adaptive per-sample budget allocation on POPE examples. Our query-aware framework dynamically scales token retention according to scene complexity. The visually redundant snow scene (left) retains only 516 of 2,880 visual tokens, whereas the cluttered baseball scene (right) dynamically allocates 860 tokens to preserve fine-grained details needed for accurate question answering.}
\label{fig:sampleallocation}
\end{figure}


Despite their success, both budget allocation and token selection in methods like HiRED and GlobalCom$^2$ rely entirely on visual saliency cues (e.g., [CLS] attention) derived solely from the image. Consequently, their pruning masks remain static across different textual queries, meaning visually inconspicuous but query-critical vision tokens can be discarded prematurely (Fig.~\ref{fig:pruningmask}). Once pruned, this vital visual evidence cannot be recovered by subsequent LLM layers, directly undermining the benefit of high-resolution encoding. While HERO~\cite{li2025hero} introduces query awareness via crop-query CLIP similarity for budget allocation, its query guidance remains restricted to the coarse crop level, leaving fine-grained token selection completely query-blind.


To overcome these limitations, we propose a training-free, top-down pruning hierarchy that infuses query guidance prior to any LLM execution. Crucially, existing methods enforce a uniform pruning budget across all images regardless of their visual information density. To address this inefficiency, our hierarchy begins at the macro \textit{instance-level} with a complexity-aware allocation: as illustrated in Fig.~\ref{fig:sampleallocation}, visually rich scenes require higher token budgets than visually redundant scenes, which we dynamically compute based on thumbnail token diversity, while maintaining a fixed benchmark average budget. At the mid \textit{crop-level}, this budget is distributed across local crops using sentence-level cross-modal similarity. Finally, at the micro \textit{token-level}, we exploit the shared LLM input space by comparing projected vision tokens directly with query-token embeddings; we normalize similarities per query token via a text-debiasing normalization and fuse the resulting scores with local and global visual attention to preserve the most critical vision tokens.


Our contributions are summarized as follows:
\begin{itemize}
    \item We introduce a pre-LLM query-scoring mechanism that computes text-debiased vision-language relevance in the shared space, combining it with local and global visual saliency to preserve the most critical vision tokens.
    \item We propose an instance-level budget allocation strategy that scales token retention ratios based on thumbnail-token diversity to improve performance.
    \item We present a plug-and-play, training-free framework that achieves a $2.0\times$ throughput speedup on LLaVA-NeXT-7B at 25\% retention, outperforming state-of-the-art baselines and surpassing the unpruned baseline on POPE.
\end{itemize}

\section{Related Work}

\subsection{High-Resolution LVLMs}

High-resolution inputs preserve critical fine-grained details, such as small objects and scene text, that are lost under aggressive downsampling. Existing high-resolution LVLMs support these inputs through three main designs. \textit{Crop-based models}, including LLaVA-NeXT~\cite{liu2024llavanext}, InternVL~\cite{chen2024internvl15}, Monkey~\cite{li2024monkey}, and LLaVA-UHD~\cite{xu2024llavauhd}, partition images into local views alongside a global thumbnail. \textit{Native-resolution models}, like Qwen2-VL~\cite{wang2024qwen2vl} and Pixtral~\cite{agrawal2024pixtral}, directly process arbitrary image scales and aspect ratios. \textit{Dual-encoder models}, such as Mini-Gemini~\cite{li2024minigemini} and LLaVA-HR~\cite{luo2025llavahr}, combine low-resolution semantic features with an auxiliary high-resolution visual stream. While effective, all three designs drastically increase vision token sequence lengths. This work focuses on reducing this token overhead for crop-based LVLMs while preserving their hierarchical thumbnail-crop structure.

\subsection{Training-Free Visual Token Pruning}

We focus on methods that reduce vision tokens at inference time without requiring model retraining. \textit{Pruning within the LLM} discards tokens after they enter the language model, leveraging cross-layer attention dynamics~\cite{chen2024fastv,zhang2025sparsevlm} or progressive layer-by-layer redundancy dropping~\cite{xing2025pyramiddrop}. Conversely, \textit{pruning before the LLM} eliminates tokens prior to executing any language block, utilizing vision-encoder visual attention~\cite{zhang2024fastervlm} or feature redundancy metrics~\cite{wen2025dart} to identify salient regions. Although effective, these existing training-free approaches rely primarily on query-agnostic visual cues or compute text relevance too late inside the heavy LLM layers. Crucially, most are formulated for a single-grid image layout and fail to model the specialized global-thumbnail-and-local-crop structure of high-resolution LVLMs.

\subsection{Visual Token Pruning for High-Resolution LVLMs}

Recent high-resolution token reduction frameworks explicitly model the hierarchical thumbnail-crop structure. HiRED~\cite{arif2025hired} utilizes global thumbnail visual attention to allocate crop budgets, ranking tokens within each crop using local visual attention. GlobalCom$^2$~\cite{liu2026globalcom2} improves on this by fusing interpolated global attention with local crop attention for fine-grained selection. HERO~\cite{li2025hero} introduces query awareness to the budget allocation stage using sentence-level cross-modal similarity, but its token-level selection remains entirely query-blind, relying on multi-layer visual attention. In contrast, while our method adopts sentence-level crop allocation, we drastically bridge the query-blind gap by introducing a pre-LLM, token-level query score and dynamically adapting the total token budget to each sample's visual complexity.


\begin{figure*}[t]
\centering
\includegraphics[width=\textwidth]{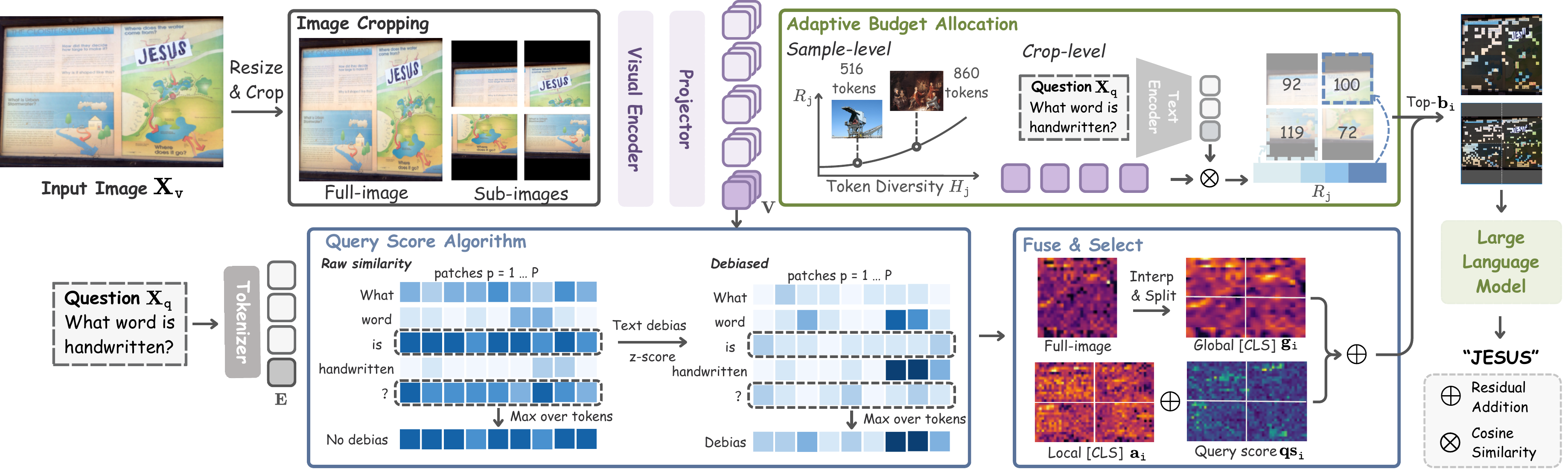}
\caption{Overview of our training-free token pruning framework for high-resolution LVLMs. An input image is processed into a full-image view and multiple sub-images, which are passed through a Visual Encoder and Projector to produce visual tokens ($V$). A text query is tokenized into embeddings ($E$). The pruning process consists of three main stages:
\textbf{(1) Adaptive Budget Allocation:} A two-level budget is computed. At the \textit{Sample-level}, a token retention ratio ($R_j$) is dynamically set for each image based on its visual token diversity ($H_j$). At the \textit{Crop-level}, this budget is distributed among the sub-images based on the query-crop relevance.
\textbf{(2) Query Score Algorithm:} A fine-grained relevance score is computed between the query embeddings ($E$) and the visual tokens ($V$). Raw cross-modal similarities are standardized using a text-debiasing z-score to produce a robust query score ($qs_i$) for each token.
\textbf{(3) Fuse \& Select:} The final token selection score is calculated by fusing the \textit{Query Score} with two visual saliency cues: a local score from the sub-image's own [CLS] attention ($a_i$) and a global score derived from the full-image view ($g_i$). The top-$b_i$ tokens are selected and fed to LLM for inference.}
\label{fig:framework}
\end{figure*}

\section{Method}
\label{sec:method}

\subsection{Framework Overview}
\label{sec:overview}


The overall pipeline of our proposed framework is illustrated in Fig.~\ref{fig:framework}. Given an input image, we split it into $n$ local crops $I_1,\dots,I_n$ and resize it to construct a global thumbnail $I_0$. A CLIP vision encoder~\cite{radford2021clip} encodes each image $I_i$ ($i=0$ for the thumbnail, $i=1,\dots,n$ for the crops) into $P=24\times24=576$ patch features. Subsequently, a projection layer maps these features into the input embedding space of the LLM. We denote the projected visual tokens as $\mathbf{V}_i=[\mathbf{v}_{i,1};\dots;\mathbf{v}_{i,P}]\in\mathbb{R}^{P\times D}$, where $D$ is the hidden size of the LLM. Because LLaVA-NeXT pads the resized image to the grid resolution, only the positions $\Omega_i\subseteq\{1,\dots,P\}$ of crop $i$ cover the image. We write $\hat{\mathbf{x}}$ for the $\ell_2$-normalized $\mathbf{x}$, with row-wise normalization for matrices, and $z(\mathbf{x})$ for standardizing a score vector by subtracting its mean over $\Omega_i$ and dividing by its standard deviation over $\Omega_i$ plus $\epsilon$. 
To achieve efficient inference, we prune visual tokens in a training-free manner after the vision encoder but prior to the LLM. This is executed via four synergistic mechanisms:
(1) Complexity-Aware Sample Budget Allocation (Sec.\ref{sec:sample}), which dynamically calibrates a custom token retention ratio for each input instance while maintaining a constant average budget across the evaluation benchmark;
(2) Query-Aware Crop Budget Allocation (Sec.\ref{sec:budget}), which adapts and extends the crop-level budgeting scheme from HERO~\cite{li2025hero};
(3) Query-Aware Token Scoring (Sec.\ref{sec:qm}), which evaluates patch-level query relevance on-the-fly prior to feeding tokens into the LLM; and
(4) Token Selection (Sec.\ref{sec:selection}), which systematically fuses query relevance with both global and local visual saliency to prune tokens under each crop's allocated budget.

\subsection{Complexity-Aware Sample Budget Allocation}
\label{sec:sample}
Existing token pruning methods apply a uniform retention ratio $R$ to all images. However, natural images vary widely in information density: an image featuring a single prominent object against a uniform background is highly redundant, whereas a cluttered scene with multiple entities contains rich, non-redundant details (see Fig.~\ref{fig:sampleallocation}). To exploit this variation, we assign each image instance $j$ its own retention ratio $R_j$, while keeping the average retention ratio across the dataset at $R$, thereby preserving the overall computational budget.

We quantify the visual complexity of instance $j$ by measuring the feature diversity of its thumbnail tokens. Let $\hat{\mathbf{v}}_{0,1}, \dots, \hat{\mathbf{v}}_{0,P}$ represent the $\ell_2$-normalized projected patch features of the global thumbnail $I_0$. We define the visual complexity metric $H_j$ as:
\begin{equation}
H_j = 1 - \Big\| \frac{1}{P} \sum_{p=1}^{P} \hat{\mathbf{v}}_{0,p} \Big\|_2^2,
\end{equation}
which is small when the thumbnail tokens point in similar directions and large when they are diverse. Crucially, because $H_j$ is computed solely on the low-resolution thumbnail $I_0$, it is computationally negligible and completely independent of the text query.

Using this complexity metric, we formulate the customized retention ratio $R_j$ for instance $j$ as:
\begin{equation}
R_j = \text{clip} \left( \kappa R \exp\left( \frac{H_j - \mu_H}{\sigma_H T_s} \right), R_{\min}, R_{\max} \right),
\end{equation}
where $T_s$ is a temperature scaling parameter, and $\mu_H$ and
$\sigma_H$ are the mean and standard deviation of visual complexity
computed over all $M$ samples in the evaluation benchmark. The scaling
coefficient $\kappa$ is chosen so that the average retention ratio
after clipping satisfies $\frac{1}{M}\sum_{j=1}^{M} R_j = R$.
The clipped bounds $[R_{\min}, R_{\max}]$ prevent extreme budget
allocations. Under this scheme, complex images are allocated more
visual tokens to preserve details, while simple images are
aggressively pruned to maximize efficiency.

\subsection{Query-Aware Crop Budget Allocation}
\label{sec:budget}

To distribute the instance-level budget $R_j$ (Sec.~\ref{sec:sample}) across different image regions, we allocate crop-level budgets based on query relevance, adapting the dual-encoder routing concept from HERO~\cite{li2025hero}. While the global thumbnail $I_0$ simply retains the $\lfloor P R_j \rfloor$ tokens with the highest last-layer [CLS] attention, local crops $\{I_1, \dots, I_n\}$ are budgeted dynamically according to their textual alignment.

Specifically, the projected CLIP [CLS] feature of crop $i$ is denoted as $\mathbf{f}_i$, and the query is encoded by the CLIP text encoder into its [EOS] embedding $\mathbf{u}$. We compute their cosine similarity to measure relevance: $c_i = \cos(\mathbf{f}_i, \mathbf{u})$.
Let $\mathbf{c}=[c_1,\dots,c_n]$. To determine the crop-level retention ratio $r_i$, we normalize these similarity scores and apply the budget transform of GlobalCom$^2$~\cite{liu2026globalcom2}:
\begin{equation}
\boldsymbol{\pi}=\mathrm{softmax}(\mathbf{c}/\tau),\qquad r_i=\min\!\big(R_j\,(1+\pi_i-\tfrac{1}{n}),\,1\big),
\end{equation}
where $\tau$ is a temperature parameter. Since $\sum_i \pi_i = 1$, the average retention ratio across all crops remains exactly $R_j$. 
This crop-level budget $b_i = \min(\lfloor Pr_i\rfloor,|\Omega_i|)$ is subsequently used to guide our fine-grained token-level pruning mechanism.

\begin{algorithm}[t]
\caption{Crop-Wise Token-Level Query Scoring}
\label{alg:qs}
\begin{algorithmic}[1]
\REQUIRE $\mathbf{V}\in\mathbb{R}^{P\times D}$, valid positions $\Omega$, query embeddings $\mathbf{E}\in\mathbb{R}^{T\times D}$
\ENSURE query scores $\mathbf{qs}\in\mathbb{R}^{|\Omega|}$
\STATE $\mathbf{S}\leftarrow\hat{\mathbf{E}}\,\hat{\mathbf{V}}^{\top}$ \hfill $\triangleright$ $T\times P$ cosine
\STATE $\mathbf{Z}[t,:]\leftarrow z(\mathbf{S}[t,:])$ for all $t$ \hfill $\triangleright$ row-wise over $\Omega$
\STATE $\mathbf{qs}\leftarrow\max_t\mathbf{Z}[t,\Omega]$
\RETURN $\mathbf{qs}$
\end{algorithmic}
\end{algorithm}

\subsection{Query-Aware Token Scoring }
\label{sec:qm}
While the crop-level relevance score $c_i$ determines the token budget of each crop, it cannot identify fine-grained, query-relevant details within a crop. Existing methods select tokens within a crop by vision-only metrics such as [CLS] attention. This visual saliency is query-agnostic, causing the model to discard visually inconspicuous yet query-critical regions (e.g., small objects, scene text, or subtle visual cues). To address this, we introduce a token-level query scoring mechanism that scores each visual token against the textual query.

Let $\mathbf{E} = [\mathbf{e}_1; \dots; \mathbf{e}_T] \in \mathbb{R}^{T \times D}$ represent the LLM input embeddings of the $T$ query tokens retrieved from the LLM's pre-trained token-embedding table. Because the vision-language projector is explicitly trained to align visual features into this exact $D$-dimensional LLM input space, the projected visual tokens $\mathbf{V}_i \in \mathbb{R}^{P \times D}$ and textual embeddings $\mathbf{E}$ reside in a shared semantic space. This alignment allows us to extract a high-fidelity cross-modal similarity signal prior to executing any LLM blocks, incurring virtually zero computational overhead. Specifically, we compute the similarity matrix $\mathbf{S}$ as:
\begin{equation}
\mathbf{S} = \hat{\mathbf{E}} \hat{\mathbf{V}}_i^{\top} \in \mathbb{R}^{T \times P},
\end{equation}
Each element $S_{t,p}=\cos(\mathbf{e}_t,\mathbf{v}_{i,p})$ is the similarity between the $t$-th query token and the $p$-th visual token of crop $i$.

Since different query tokens exhibit highly varying baseline similarity scales (e.g., high-frequency punctuation or function tokens may yield moderate similarity across all visual tokens), raw similarity scores are not directly comparable across tokens. We resolve this by applying row-wise standardization over $\Omega_i$. Let $\mathbf{s}_t$ denote row $t$ of $\mathbf{S}$. The standardized score $z(\mathbf{s}_t)_p$ measures how much visual token $p$ stands out for query token $t$ relative to the other valid tokens in the crop, removing baseline differences between query tokens.

Finally, we define the query score ($\mathrm{qs}_{i,p}$) of visual token $p$ as its maximum relative alignment across all query tokens:
\begin{equation}
\mathrm{qs}_{i,p} = \max_{t} z(\mathbf{s}_t)_p, \qquad p \in \Omega_i.
\end{equation}
Under this formulation, a visual token receives a high query score if it is distinctively relevant to even a single query token (e.g., a specific noun in the prompt), regardless of its visual prominence. Algorithm~\ref{alg:qs} summarizes the crop-wise query scoring, and the resulting scores are then fused with visual saliency for token selection.

\begin{table*}[!t]
\centering
\caption{Comparison with existing methods on LLaVA-NeXT-7B. The retention ratio $R$ is defined with respect to the 2,880 token positions including padding. Avg.\ is the mean score relative to the unpruned model. Baseline results without a mark are taken from GlobalCom$^2$~\cite{liu2026globalcom2}; $^*$ denotes results obtained by running the released code. The best and second-best results among pruning methods are in bold and underlined.}
\label{tab:main}
\setlength{\tabcolsep}{5pt}
\begin{tabular}{lccccccccc}
\toprule
Method & VizWiz & GQA & SQA-I & TextVQA & MME-P & MMB & AI2D & POPE & Avg.\ (\%) \\
\midrule
\multicolumn{10}{c}{\textit{Retention ratio $R=25\%$}} \\
\midrule
LLaVA-NeXT-7B & 60.6 & 64.3 & 70.3 & 61.4 & 1519.3 & 67.1 & 65.3 & 87.6 & 100.0 \\
\midrule
FastV~\cite{chen2024fastv} & 54.2 & 60.4 & \textbf{68.8} & 58.4 & 1477.3 & 65.6 & 63.9$^*$ & 83.1 & 95.5 \\
SparseVLM~\cite{zhang2025sparsevlm} & 56.0 & 59.9 & 67.5 & 58.3 & 1465.9 & 65.6 & \underline{64.2}$^*$ & 85.0 & 95.8 \\
PyramidDrop~\cite{xing2025pyramiddrop} & 56.8 & 60.3 & \underline{68.5} & 59.8 & 1473.7 & 65.6 & 63.9$^*$ & 85.5 & 96.6 \\
FasterVLM~\cite{zhang2024fastervlm} & 55.4 & 61.3 & 67.1 & 58.8 & 1454.6 & \textbf{66.0} & 62.6$^*$ & 87.2 & 95.9 \\
HiRED~\cite{arif2025hired} & 58.7$^*$ & 61.9$^*$ & 68.4$^*$ & \underline{60.0}$^*$ & 1480.5$^*$ & 65.5$^*$ & 61.5$^*$ & 86.9$^*$ & 97.1 \\
HERO~\cite{li2025hero} & \textbf{59.8}$^*$ & 62.6$^*$ & 67.7$^*$ & 58.9$^*$ & 1491.6$^*$ & 65.2$^*$ & \textbf{64.5}$^*$ & 87.7$^*$ & \underline{97.8} \\
GlobalCom$^2$~\cite{liu2026globalcom2} & 55.7 & 61.5 & 68.1 & \textbf{60.9} & 1493.5 & \underline{65.9} & 63.2$^*$ & 87.6 & 97.1 \\
Ours (fixed $R$) & \underline{58.9} & \textbf{63.1} & 68.0 & 59.3 & \underline{1501.4} & 65.1 & 63.4 & \textbf{88.5} & \underline{97.8} \\
Ours & 58.7 & \underline{63.0} & 67.9 & 59.4 & \textbf{1514.1} & 65.6 & 63.7 & \underline{88.3} & \textbf{98.0} \\
\midrule
\multicolumn{10}{c}{\textit{Retention ratio $R=10\%$}} \\
\midrule
LLaVA-NeXT-7B & 60.6 & 64.3 & 70.3 & 61.4 & 1519.3 & 67.1 & 65.3 & 87.6 & 100.0 \\
\midrule
FastV~\cite{chen2024fastv} & 53.1 & 55.9 & 68.1 & 55.7 & 1282.9 & 61.6 & \underline{62.2}$^*$ & 71.7 & 89.4 \\
SparseVLM~\cite{zhang2025sparsevlm} & 52.1 & 56.3 & \underline{68.5} & 53.9 & 1334.2 & 60.0 & 61.7$^*$ & 80.1 & 90.2 \\
PyramidDrop~\cite{xing2025pyramiddrop} & 54.4 & 54.5 & 67.7 & 54.4 & 1262.1 & 59.0 & 60.8$^*$ & 77.6 & 89.0 \\
FasterVLM~\cite{zhang2024fastervlm} & 52.6 & 56.9 & 66.5 & 56.5 & 1359.2 & 61.6 & 61.8$^*$ & 83.6 & 91.7 \\
HiRED~\cite{arif2025hired} & \textbf{58.3}$^*$ & 58.8$^*$ & 67.2$^*$ & \textbf{58.5}$^*$ & 1365.7$^*$ & \textbf{63.6}$^*$ & 60.3$^*$ & 85.1$^*$ & 94.1 \\
HERO~\cite{li2025hero} & 56.5$^*$ & 58.7$^*$ & 67.4$^*$ & 56.2$^*$ & \textbf{1436.6}$^*$ & \underline{62.8}$^*$ & \textbf{62.5}$^*$ & \underline{85.2}$^*$ & 94.1 \\
GlobalCom$^2$~\cite{liu2026globalcom2} & 54.6 & 57.1 & \textbf{68.7} & \underline{58.4} & 1365.5 & 61.8 & 61.4$^*$ & 83.8 & 92.9 \\
Ours (fixed $R$) & 56.5 & \underline{59.3} & 67.5 & 58.0 & \underline{1433.1} & 61.5 & 62.0 & \textbf{87.8} & \underline{94.6} \\
Ours & \underline{56.6} & \textbf{59.4} & 67.7 & 58.1 & 1428.3 & 61.9 & \underline{62.2} & \textbf{87.8} & \textbf{94.8} \\
\bottomrule
\end{tabular}
\end{table*}

\subsection{Token Selection}
\label{sec:selection}

Given the allocated crop budget $b_i$, we select the optimal subset of visual tokens for each crop by fusing a global context score with a query-aware local score. Following GlobalCom$^2$, the global score $g_{i,p}$ is obtained by bilinearly interpolating the last-layer [CLS] attention map of the global thumbnail $I_0$ to the resolution of the unpadded high-resolution feature map, specifically extracting the spatial region corresponding to crop $i$. This term reflects the relative importance of visual token $p$ in the context of the entire image. 
The local score $l_{i,p}$ balances the crop's intrinsic visual saliency with its query relevance:
\begin{equation}
l_{i,p} = \beta \, z(\mathbf{a}_i)_p + (1-\beta) \, z(\mathbf{qs}_i)_p,
\end{equation}
where $\mathbf{a}_i$ represents the last-layer [CLS] attention map computed on the crop itself and $\mathbf{qs}_i$ is the token-level query score. 
The final selection score $s_{i,p}$ fuses the global and local terms after performing $\ell_2$ normalization over the set of valid tokens:
\begin{equation}
s_{i,p} = \alpha \, \frac{g_{i,p}}{\|\mathbf{g}_i\|_2} + (1-\alpha) \, \frac{l_{i,p}}{\|\mathbf{l}_i\|_2},
\end{equation}
where $\alpha \in [0, 1]$ is a hyperparameter balancing global and local cues. Finally, the $b_i$ visual tokens yielding the highest final scores are preserved:
\begin{equation}
\mathcal{K}_i = \operatorname*{top\text{-}} b_i \left( \{s_{i,p}\}_{p \in \Omega_i} \right).
\end{equation}
The visual tokens sent to the LLM consist of the retained thumbnail tokens together with $\mathcal{K}_1\cup\dots\cup\mathcal{K}_n$.

\section{Experiments}
\label{sec:exp}

\subsection{Experimental Setting}
\label{sec:setting}

We evaluate our proposed method on LLaVA-NeXT-7B~\cite{liu2024llavanext} using a single NVIDIA RTX 4090 GPU. We compare against FastV~\cite{chen2024fastv}, SparseVLM~\cite{zhang2025sparsevlm}, PyramidDrop~\cite{xing2025pyramiddrop}, FasterVLM~\cite{zhang2024fastervlm}, HiRED~\cite{arif2025hired}, HERO~\cite{li2025hero}, and GlobalCom$^2$~\cite{liu2026globalcom2} at various token retention ratios $R$. Our default hyperparameters follow the configuration ($T_s = 10$, $R_{\min} = 0.72R$, $R_{\max} = 2R$, $\tau = 0.01$, $\alpha = 0.7$, and $\beta = 0.5$).

We evaluate on eight benchmarks: VizWiz~\cite{gurari2018vizwiz}, GQA~\cite{hudson2019gqa}, ScienceQA-IMG~\cite{lu2022scienceqa}, TextVQA~\cite{singh2019textvqa}, MME~\cite{fu2023mme}, MMBench~\cite{liu2024mmbench}, AI2D~\cite{kembhavi2016ai2d}, and POPE~\cite{li2023pope}. We report accuracy on all benchmarks except MME, for which we report the perception score, and POPE, for which we report the F1 score averaged over its three subsets. Avg.\ is the mean of the per-benchmark scores normalized by the unpruned model. $R$ is defined with respect to the 2,880 token positions including padding, following HiRED~\cite{arif2025hired} and HERO~\cite{li2025hero}.

\subsection{Main Results}
\label{sec:main}

As shown in Table~\ref{tab:main}, our method achieves the highest average performance on LLaVA-NeXT-7B at both $R=0.25$ and $R=0.10$, achieving relative performance scores of $98.0$ and $94.8$, respectively, with the unpruned baseline normalized to $100.0$. This outperforms the strongest baseline at $R=0.25$, HERO ($97.8$), and the strongest baselines at $R=0.10$, HiRED and HERO (both $94.1$). Introducing complexity-aware budget allocation further improves average performance by 0.2 points at both ratios ($98.0$ vs. $97.8$ and $94.8$ vs. $94.6$) without increasing the average vision token count. Notably, our method surpasses the unpruned model on POPE at both retention ratios ($88.3$ and $87.8$ vs. $87.6$). Since POPE evaluates object hallucination, these gains demonstrate that pruning query-irrelevant vision tokens effectively mitigates model hallucination.

\subsection{Ablation Study and Analysis}
\label{sec:ablation}

\begin{table}[t]
\centering
\caption{Module ablation on LLaVA-NeXT-7B at $R=25\%$.}
\label{tab:module}
\begin{tabular}{cc|rr}
\toprule[1pt]
Sample alloc. & Query score & MME-P & AI2D \\
\midrule
 & \checkmark & 1501.4 & 63.4 \\
\checkmark & & 1506.5 & 63.6 \\
\checkmark & \checkmark & \textbf{1514.1} & \textbf{63.7} \\
\bottomrule[1pt]
\end{tabular}
\end{table}

\paragraph{Module Ablation} As shown in Table~\ref{tab:module}, removing either Complexity-Aware Allocation or Query Score mechanism from the framework degrades performance. Replacing the adaptive per-sample ratio $R_j$ with a fixed ratio $R$ drops MME-P from 1514.1 to 1501.4 and AI2D from 63.7 to 63.4. Removing the query score ($\beta=1$) drops MME-P from 1514.1 to 1506.5 and AI2D from 63.7 to 63.6, demonstrating that both complexity-aware budget scaling and query-aware vision token scoring are critical to achieving the overall accuracy gains.

\begin{table}[t]
\centering
\caption{Ablation on $\alpha$ at $R=25\%$ ($\beta=0.5$). $\alpha=1$ uses only the global score; the default is underlined.}
\label{tab:alpha}
\begin{tabular}{l|ccccccc}
\toprule[1pt]
$\alpha$ & 0.0 & 0.1 & 0.3 & 0.5 & \underline{0.7} & 0.9 & 1.0 \\
\midrule
MMB & 65.4 & 65.3 & 65.3 & 65.2 & \textbf{65.6} & 65.0 & 65.1 \\
POPE & 88.1 & 88.1 & 88.2 & 88.2 & \textbf{88.3} & \textbf{88.3} & 87.7 \\
\bottomrule[1pt]
\end{tabular}
\end{table}

\paragraph{Influence of Global-Local Balance $\alpha$} Table~\ref{tab:alpha} evaluates the balance weight $\alpha$ between global and local scores. Because the global score is interpolated from the thumbnail's [CLS] attention, it varies smoothly across space; a larger $\alpha$ favors contiguous vision token regions, preserving spatial semantic integrity. We find $\alpha=0.7$ yields the best performance on MMBench. However, relying solely on global cues ($\alpha=1.0$) drops POPE from 88.3 to 87.7, showing that the local score is needed for precise object verification.

\begin{table}[t]
\centering
\caption{Ablation on $\beta$ at $R=25\%$ ($\alpha=0.7$). $\beta=0$ uses only the query score as the local score; the default is underlined. $\beta=1$ removes the query score.}
\label{tab:beta}
\setlength{\tabcolsep}{4pt}
\begin{tabular}{l|ccccccc}
\toprule[1pt]
$\beta$ & 0.0 & 0.1 & 0.3 & \underline{0.5} & 0.7 & 0.9 & 1.0 \\
\midrule
TextVQA & 59.1 & 59.2 & \textbf{59.4} & \textbf{59.4} & \textbf{59.4} & 59.2 & 59.3 \\
MMB & 65.3 & 65.0 & 65.2 & \textbf{65.6} & 65.5 & 65.3 & 65.3\\
\bottomrule[1pt]
\end{tabular}
\end{table}


\paragraph{Influence of Query-Saliency Balance $\beta$} Table~\ref{tab:beta} varies the weight $\beta$ balancing crop [CLS] attention against the query score. Setting $\beta=0.5$ yields the optimal trade-off. While the text-matching query score accurately pinpoints question-focused vision tokens, they can be spatially scattered as they lack visual saliency context. Weighting both terms equally allows the visual saliency and cross-modal query scores to complement each other effectively.

\begin{table}[t]
\centering
\caption{Inference efficiency of LLaVA-NeXT-7B on POPE at $R=25\%$. TFLOPs: LLM prefill FLOPs. Memory: peak GPU memory. Throughput: questions per second. Performance: average relative score in Table~\ref{tab:main}.}
\label{tab:efficiency}
\setlength{\tabcolsep}{3pt}
\resizebox{\columnwidth}{!}{%
\begin{tabular}{l|cccc}
\toprule[1pt]
Method  & TFLOPs$\downarrow$ & Mem.$\downarrow$ (GB) & Thr.$\uparrow$ (q/s) & Perf.$\uparrow$ \\
\midrule
LLaVA-NeXT-7B  & 32.78 & 16.55 & 3.15 (1.0$\times$) & 100.0\% \\
\midrule
HERO~\cite{li2025hero} & 10.88 & 15.44 & 4.87 (1.5$\times$) & 97.8\% \\
Ours (fixed $R$) & 10.88 & 14.55 & 6.34 (2.0$\times$) & 97.8\% \\
Ours& 10.88 & 14.84 & 6.35 (2.0$\times$) & \textbf{98.0\%} \\
\bottomrule[1pt]
\end{tabular}}
\end{table}

\subsection{Efficiency Analysis}
\label{sec:efficiency}

Table~\ref{tab:efficiency} compares inference efficiency on the POPE benchmark (9,000 queries) with a batch size of 1. At $R=25\%$, our framework maintains the same target vision token count and prefill FLOPs as HERO. However, our method achieves a higher throughput than HERO, while simultaneously securing the highest average performance.

\section{Conclusion}
\label{sec:conclusion}

In this paper, we propose a training-free, hierarchical vision token pruning framework for high-resolution LVLMs. Our top-down pipeline dynamically scales visual budgets per sample based on visual complexity (instance-level), distributes this budget across crops using cross-modal similarity (crop-level), and selects tokens within each crop by combining visual saliency with a text-debiased query score computed in the shared input space (token-level). Extensive evaluations on LLaVA-NeXT-7B demonstrate that our plug-and-play method achieves state-of-the-art performance across diverse benchmarks at both 25\% and 10\% token retention. Notably, at 25\% retention, our framework delivers a 2.0$\times$ inference throughput speedup while even surpassing the unpruned baseline on POPE, highlighting the immense potential of query-aware pruning for efficient multimodal inference.

\FloatBarrier

\bibliographystyle{IEEEtran}
\bibliography{refs,bench_refs}

\end{document}